\pdfoutput=1

\documentclass[11pt]{article}
\usepackage{authblk}

\usepackage[]{ACL2023}

\usepackage{times}
\usepackage{latexsym}
\usepackage{graphicx}
\usepackage{tabularx}
\usepackage[T1]{fontenc}

\usepackage[utf8]{inputenc}

\usepackage{microtype}

\usepackage{inconsolata}

\usepackage{multirow}

\usepackage{enumitem} 

\usepackage{booktabs}

\title{Efficient Data Generation for Source-grounded \\ Information-seeking Dialogs: A Use Case for Meeting Transcripts}

\author{
 \textbf{Lotem Golany\textsuperscript{*,1,3}},
 \textbf{Filippo Galgani\textsuperscript{*,1}},
 \textbf{Maya Mamo\textsuperscript{*,1}},
 \textbf{Nimrod Parasol\textsuperscript{1}},
\\
 \textbf{Omer Vandsburger\textsuperscript{1}},
 \textbf{Nadav Bar\textsuperscript{1}},
 \textbf{Ido Dagan\textsuperscript{1,2}}
\\
 \textsuperscript{1}Google Research,
 \textsuperscript{2}Bar-Ilan University
\\
 \small{
     \textsuperscript{3}\textbf{Corresponding author:} \href{mailto:mlotem@google.com}{mlotem@google.com}
 }
}

\begin{document}
\maketitle

\begin{abstract}
Automating data generation with Large Language Models (LLMs) has become increasingly popular. In this work, we investigate the feasibility and effectiveness of LLM-based data generation in the challenging setting of source-grounded information-seeking dialogs, with response attribution, over long documents. Our source texts consist of long and noisy meeting transcripts, adding to the task complexity.
Since automating attribution remains difficult, we propose a semi-automatic approach: dialog queries and responses are generated with LLMs, followed by human verification and identification of attribution spans.
Using this approach, we created MISeD -- Meeting Information Seeking Dialogs dataset -- a dataset of information-seeking dialogs focused on meeting transcripts. 
Models finetuned with MISeD demonstrate superior performance compared to off-the-shelf models, even those of larger size. Finetuning on MISeD gives comparable response generation quality to finetuning on fully manual data, while improving attribution quality and reducing time and effort.
\end{abstract}

\def\thefootnote{*}\footnotetext{These authors contributed equally.}\def\thefootnote{\arabic{footnote}}

\section{Introduction}
Source-grounded information-seeking dialogs allow users to efficiently navigate within a given knowledge source and extract information of interest.
In this conversational setting, a user interacts with an agent over multiple rounds of queries and responses regarding the source text (\citealp{coqa:2019}, \citealp{gopalakrishnan19_interspeech}, \citealp{doc2dial:2020}). To train effective agent models, quality dialog datasets are essential. 

\begin{figure}[ht]
    \centering
    \includegraphics[width=0.9\columnwidth]{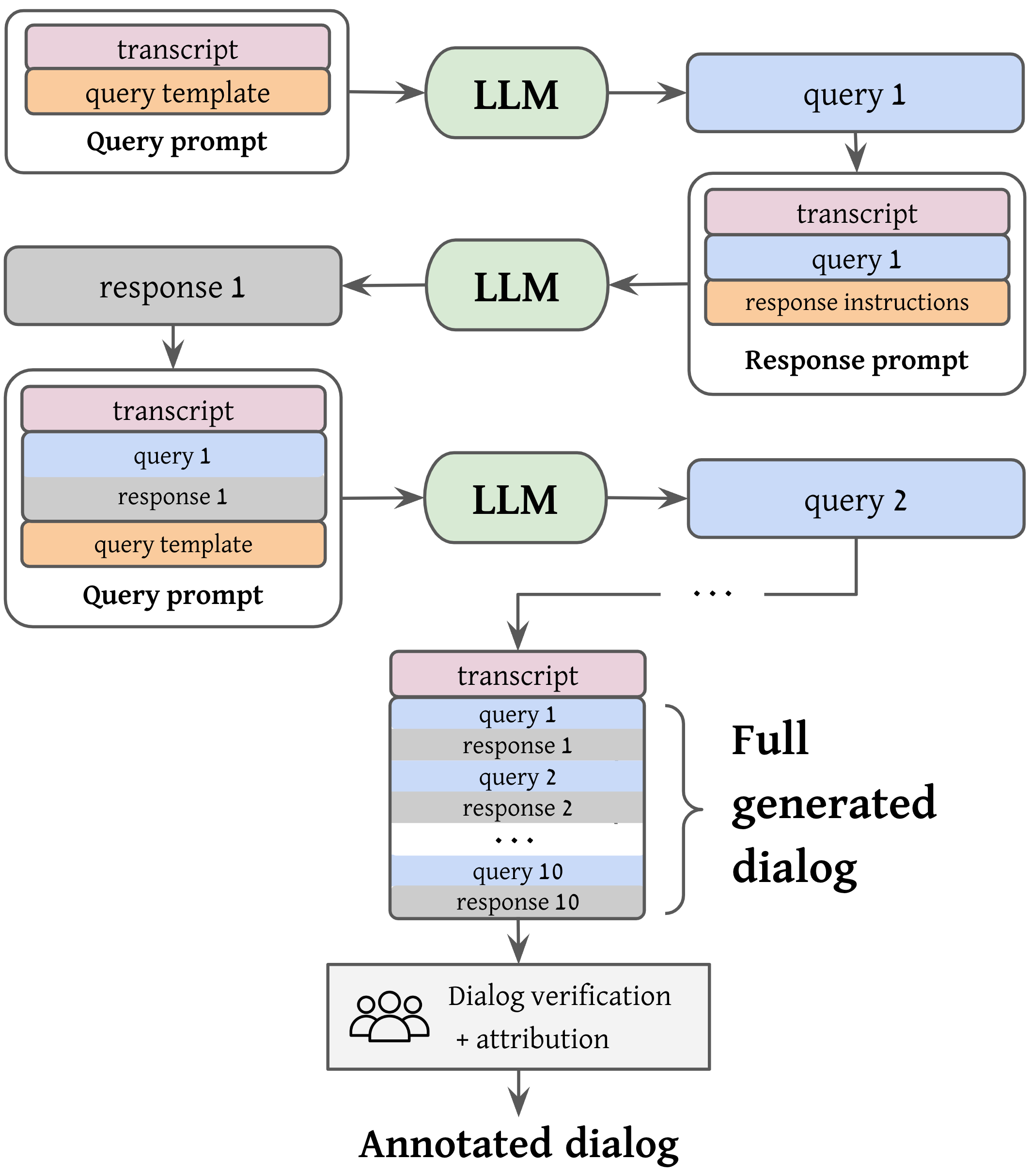}
    \caption{Iterative dialog generation flow. In each turn, a query prompt guides the LLM to generate a user query given the transcript, the accumulated dialog history, and a query template. Then, a response prompt, accompanied by the full context so far, generates the agent response. Iterating this automatic process yields a full dialog, which is then validated by annotators, who further augment it with response attributions.}
    \label{fig:dialog-generation}
\end{figure}

The prominent (manual) technique for creating dialog datasets is the Wizard-of-Oz (WOZ) method (\citealp{woz:1984}, \citealp{multiwoz:2018}), in which two human annotators collaboratively produce dialogs: one annotator acts as the user, asking questions about a hidden text, while the other annotator acts as the agent, using the source text to provide answers. The fully manual WOZ methodology is often time-consuming and can result in answers that vary in quality across annotators.

Previous work has explored the application of LLM-based dialog generation in domains like everyday conversations \citep{chen2023places} or task-oriented dialogs \citep{sun2021adding, mehri2022lad}.
In this work, we investigate potential automation of the WOZ process to create source-grounded information-seeking dialogs. While LLMs can generate dialog content, attribution generation currently requires human involvement for reasonable quality. Therefore we propose a semi-automatic approach: using prompts to guide LLM generation of queries and responses, followed by manual attribution and validation of response correctness (Figure \ref{fig:dialog-generation}). By comparing our semi-automatic approach to the traditional, fully-manual WOZ method, we demonstrate that a model trained on our data achieves comparable response generation quality while improving attribution and reducing costs.

We apply our methodology to create MISeD -- the first dataset for information-seeking dialogs over meeting transcripts, supporting the use-case of users catching up on meetings they have missed. 
Meeting transcripts present additional challenges as they often contain disfluencies, interruptions, and off-topic comments—challenges that are also representative of a broad range of real-world conversational settings. Existing datasets in the meeting domain provide summarization and question-answering data (e.g., \citealt{qmsum:21}, \citealt{meetingqa:2023}), but none supports multi-turn dialogs over meeting content. 
The dialog setup introduces additional complexity, as each query and response must consider the shared dialog state (e.g., previously shared information) in addition to the transcript, to maintain a coherent dialog flow.

We further present an evaluation framework and few baseline models, whose evaluation demonstrates the benefit of training with our MISeD data.

In summary, our main contributions are:
(1) we explore the feasibility of using LLMs for data generation in source-grounded information-seeking dialogs, demonstrating improved overall quality and efficiency compared to a fully manual approach;
(2) we release the MISeD dataset\footnote{We make our datasets publicly available at https://github.com/google-research-datasets/MISeD.} -- the first dialog dataset over meeting transcripts, consisting of verified source-grounded dialogs with transcript attribution, as well as a corresponding smaller fully manual (WOZ) dataset;
(3) we introduce baseline models, an evaluation procedure, and baseline results for meeting-grounded dialog tasks, through which we assess our proposed approach and the MISeD dataset.
\section{Related Work}
This section provides background on the two research areas which our work bridges: source-grounded information-seeking dialogs (\S\ref{info-seeking-related-work}) and summarization and question answering (Q\&A) over meeting transcripts (\S\ref{meetings-related-work}).

\subsection{Source-grounded Information-seeking Dialogs} \label{info-seeking-related-work}
Source-grounded information-seeking dialogs are multi-turn conversational interactions, where users seek information from a given source text.
For each user query, the agent model provides a response, with supporting references (attributions). 

Existing datasets for this task address different types of knowledge sources. Some retrieve answers from large textual corpora (\citealp{wizardofwikipedia:2019}, \citealp{doqa:2020}, \citealp{QReCC:2021}, \citealp{topiocqa:2022}), others rely on short text passages (\citealp{quac:2018}, \citealp{sharc:2018}, \citealp{coqa:2019}, \citealp{hybridialogue:2022}), long informative conversations (\citealp{qaconv:2022}), or news articles (\citealp{newsdialogues:2023}).
Our work creates a dialog dataset over the source of long meeting transcripts.

The prominent approach to creating information-seeking dialog datasets is Wizard-of-Oz (WOZ) (\citealp{woz:1984}). In this setup, two annotators roleplay as a user and an agent: the user annotator asks questions about the given hidden source, while the agent annotator, who can access the source, provides corresponding answers.

Building on recent work to automate dialog data generation with LLMs \citep{lin-etal-2022-knowledge, wu-etal-2022-dg2, synthetic:2023,Zheng2022AugESCDA,sun2021adding,mehri2022lad,chen2023places,li2022controllable,Chen2022WeaklySD}, we focus on source-grounded information-seeking dialog creation through a combination of LLM-based generation and human verification and attribution.

\subsection{Summarization and Q\&A over Meeting Transcripts} \label{meetings-related-work}

Meeting transcripts pose unique challenges due to their unstructured and lengthy nature, and potential speech recognition errors.
Existing datasets for inquiring meeting content are limited to single-turn settings, focusing on summarization and Q\&A.
 Examples include AMI (\citealp{ami:2005}) and ICSI (\citealp{icsi:2003}) for meeting summarization, ELITR (\citealp{nedoluzhko-etal-2022-elitr}) and MeetingBank (\citealp{meetingbank:2023}) for meeting minuting,
MUG (\citealp{zhang2023mug}) with summaries and  additional annotations, and ExplainMeetSum (\citealp{kim-etal-2023-explainmeetsum}) that also incorporates attributions.

MeeQA \citep{apel2023meeqa} and MeetingQA \citep{meetingqa:2023} are additional Q\&A datasets based on questions asked by the participants during the meeting itself.
Differently from this body of work, we focus on a free dialog setup, where queries are not limited, and can build upon each other (e.g., follow-up questions). To our knowledge, there is no existing dataset for information-seeking dialogs over meeting transcripts.

Our work leverages QMSum (\citealp{qmsum:21}), a widely used dataset for single-turn query-based summarization over meeting transcripts, where summaries are also supported by attribution text spans.
QMSum's creation involved annotators generating queries based on a predefined schema, including both general queries addressing the entire meeting and specific queries targeting identified topics and participants.
In our work, we modify the QMSum schema to suit LLM prompts (rather than human annotators) and extend it to generate multi-turn dialogs rather than standalone question-answer pairs. After analyzing user data collected in our system prototype, we incorporated additional query types that were not covered by QMSum, such as unanswerable queries, yes/no queries, and follow-up, context-dependent dialog queries.
Section \ref{results-qmsum} compares our dataset with QMSum for model training, and Table \ref{mised_vs_qmsum_examples} (Appendix) shows examples of MISeD data compared to QMSum data.

\section{Source-grounded Information-seeking Dialogs: Task Definition}\label{sectiontask}
\begin{figure}[t]
    \centering
    \includegraphics[width=\columnwidth]{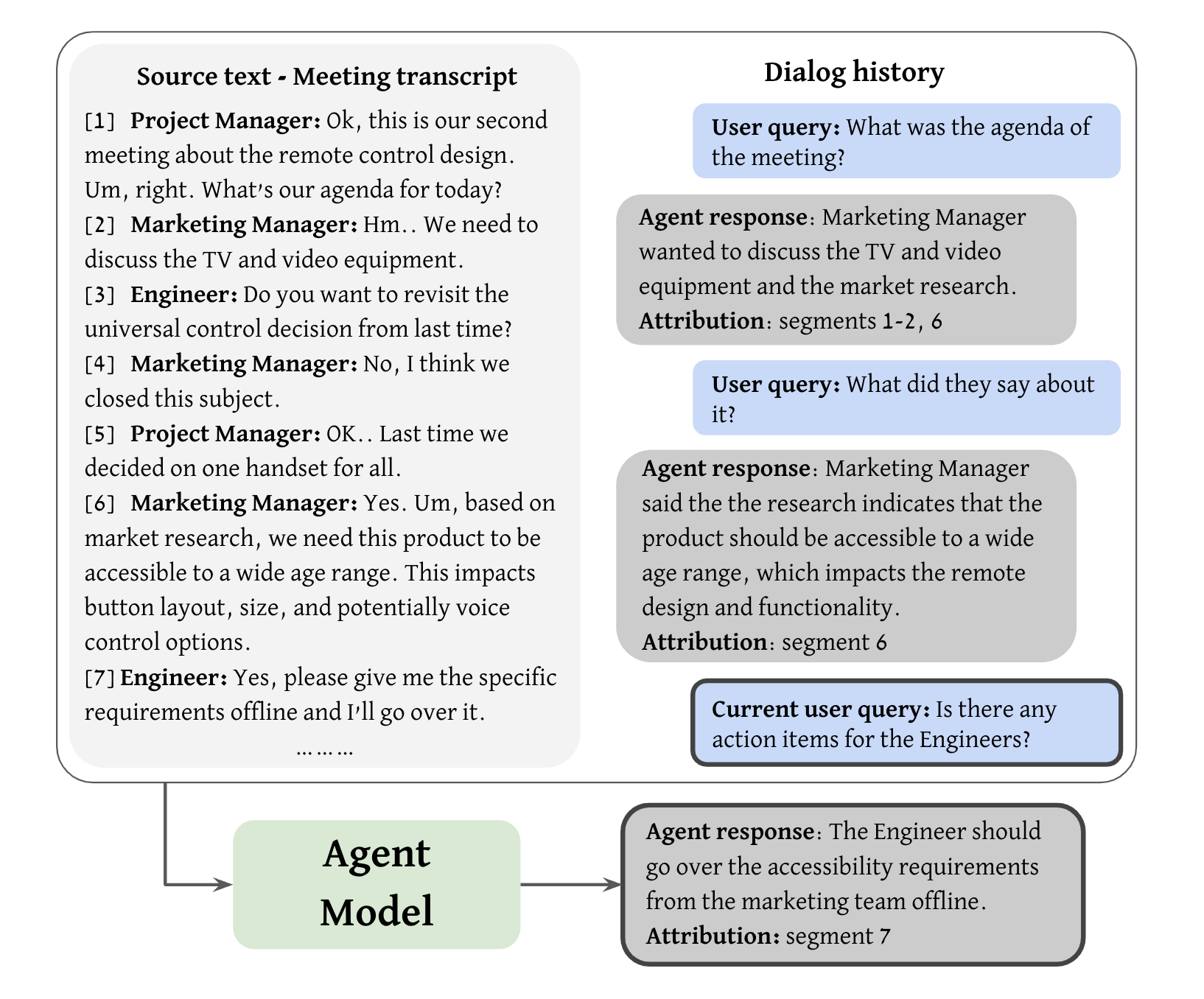}
    \caption{An illustration of the agent model task. The agent receives the source text, dialog history, and the current user query. It then generates a corresponding response along with supporting attributions in the source text. Each attribution is a sequence of consecutive transcript segments.}
    \label{fig:task-description}
\end{figure}

As described above, we focus on generating data to train agent models for the task of source-grounded information-seeking dialogs.
The agent model task (demonstrated in Figure \ref{fig:task-description}) is defined as follows: at each dialog turn the user issues a query about the source text. The agent model receives the source text, the preceding dialog history, and the current query, and is tasked with generating a response and providing the supporting attributions. Each attribution is a consecutive span in the source text.

\section{Dataset Creation Methodology} \label{data-creation-methodology}
This section details our semi-automatic dataset creation methodology, as applied to meeting transcripts (Figure \ref{fig:dialog-generation}). We first automatically generate dialogs using a pre-trained LLM (\S\ref{automated-dialog-generation}), simulating the typical WOZ process. Then we employ human annotators to generate response attributions, a task that currently necessitates human expertise. As part of the attribution process, annotators also verify the accuracy and validity of the generated queries and responses (\S\ref{dialog-validation}).

\subsection{Automatic Dialog Generation} \label{automated-dialog-generation}
We generate dialog turns iteratively, with each turn consisting of a user query and an agent response.
These are generated via targeted LLM prompts, designed for queries (\S\ref{Request prompts}) and responses (\S\ref{Response prompts}). 

\subsubsection{User query prompts}  
\label{Request prompts}
In each turn, a user query is generated by a "query prompt" (Table \ref{query_prompt} in Appendix \ref{appendix-a}). This prompt incorporates the transcript, the dialog history, and a templated instruction randomly selected from a pool of query instructions (Table \ref{full_query_templates_list} in Appendix \ref{appendix-a}), designed to include various query types.

Starting from the QMSum schema (\S\ref{meetings-related-work}), we 
create a corresponding set of query templates adapted for guiding LLMs.
We include both \textit{General} queries for overall meeting themes, and \textit{Specific} queries which focus on particular topics or individuals. We also expand the schema and include \textit{Yes-no} questions, an \textit{Unanswerable} variant, to generate queries that cannot be answered from the transcript, and \textit{Context-dependent} queries that rely on the existing dialog for context.

While our set of templates does not cover all possible query types in the meeting domain, we suggest that it represents a sufficiently broad range to yield a useful dataset, as assessed in our experiments (\S\ref{models-and-results}). Further, we propose that crafting domain-specific query template schemas can be similarly feasible and effective in other domains.

\subsubsection{Agent response prompt} \label{Response prompts}
After generating the user query, we provide the LLM with a prompt that includes the meeting transcript, dialog history, the current query, and an instruction on how to generate the agent response (Table \ref{table:response_prompt}). The instruction guides the LLM to generate an answer that aligns with certain length and format guidelines, and that is grounded in the transcript. It further emphasizes ethical and unbiased communication by instructing the model to use neutral language and avoid direct quotes.

\subsection{Dialog Annotation and Validation} \label{dialog-validation}
Following automatic dialog generation, trained annotators assessed the generated queries and responses while identifying supporting attribution spans within the source text. We first employed 3 professional annotators (details in Appendix \ref{appendix-annotations}) and conducted a pilot study, measuring Cohen’s Kappa agreement between each pair of annotators, over 525 dialog turns. Agreement scores are presented in Table \ref{agreement}.

\begin{table*}
\centering
\small
\begin{tabular}{l >{\centering\arraybackslash}p{1.5cm} >{\centering\arraybackslash}p{1.5cm} >{\centering\arraybackslash}p{1.5cm}}
\textbf{Assessment Question} & \textbf{Pair A} & \textbf{Pair B} & \textbf{Pair C} \\
\toprule 
“Is the question understandable and makes sense?” & 1.0 & 1.0 & 1.0 \\
“Does the meeting transcript contain the right answer?” & 1.0 & 0.93 & 0.93 \\ 
“Is the provided answer correct based on the meeting transcript?” & 0.86 & 0.90 & 0.80 \\
\bottomrule 
\end{tabular}
\caption{Cohen's Kappa agreement between each pair of annotators, for each assessment question. Annotators were asked to edit the model response, if needed, in case they replied “yes” to the first two questions.}
\label{agreement}
\end{table*}

The high level of agreement observed confirmed the annotation task's clarity and criteria consistency, allowing us to proceed with a single annotator annotating each dialog in the main phase.

\subsubsection{Query assessment}\label{sec:assesing_query}
Annotators verify query validity by answering the question: "\textit{Is the question understandable and makes sense?}". If a query is marked as invalid, it is removed from the dialog. We also remove all subsequent dialog turns to avoid cascading errors. Valid queries are then annotated with metadata, including their type (`general', `specific', or `yes/no') and whether they depend on prior turns (`context-dependent' or not). Context-Dependent query templates specifically target context-dependent queries, so we expect these turns to be context dependent, but annotators still check for context-dependency to confirm that the query does depend on previous turns.

\subsubsection{Response annotation: attribution, validation, and editing}
\label{sec_dataset_annotation}
Due to the current limitations of LLM-based attribution detection (which we also observed in \S \ref{result_attribution_subsection} and is reported in the literature e.g. \citep{gao2023enabling}), we opted for fully manual attribution annotation. For each generated response, annotators identify supporting text spans in the transcript while also verifying that the response is factually correct and editing the response as necessary to ensure its accuracy (more details in Appendix \ref{appendix-annotations}).
If a query cannot be answered from the transcript, it is marked as `unanswerable' and annotators ensure the response accurately conveys the lack of information.

\section{Datasets}
This section describes the MISeD dataset (\S \ref{dataset-description}), constructed using our semi-authomatic method (\S \ref{data-creation-methodology}). We also present a smaller independent  dataset created using the fully manual Wizard-of-Oz methodology (\S \ref{wizard-of-oz}), created to assess whether MISeD data is comparable to human-generated data.
Finally, we present a manual assessment of MISeD response quality (\S \ref{quality-assessment}).

\subsection{MISeD Dataset}\label{dataset-description}

\paragraph{Meeting sources}
Our data creation method was applied on transcripts from the QMSum meeting corpus.
We used 225 meetings across three domains:
134 Product Meetings (AMI; \citealp{ami:2005}), 58 Academic Meetings (ICSI; \citealp{icsi:2003}), and 33 public Parliamentary Committee Meetings. Figure \ref{fig:trancript_and_attributions} presents the transcript lengths of meetings in MISeD.

\paragraph{LLM}
We used the public Gemini Pro model \citep{team2023gemini} to automatically generate dialogs based on the meeting transcripts, as described in Section \ref{automated-dialog-generation}. 

\subsubsection{Dataset structure}
Each dataset instance includes a single dialog about a specific meeting transcript, containing up to ten query-response turns with associated metadata (\S\ref{sec:assesing_query}). When relevant, a response is accompanied by a set of attributing transcript spans.\footnote{Each span is a sequence of consecutive transcript segments.} In some cases, the response inherently lacks attributions, mostly for `unanswerable' queries.

For training and evaluating an agent model, each dialog is divided into task instances. Each such instance represents a single current query, incorporating its preceding dialog history along with the corresponding target response and attributions.

\subsubsection{Dataset statistics}
Statistics of the final MISeD dataset are presented in Table \ref{MISeD-dataset-statistis}.

\begin{table}
\centering
\small
\begin{tabular}{l >{\centering\arraybackslash}p{2cm}}
\textbf{Overall MISeD Statistics} & \\
\toprule 
\# meetings & 225 \\
\# dialogs & 432\\ 
\# query-response pairs & 4161 \\ 
\cmidrule{1-2}
\begin{tabular}{ll}
    & \begin{tabular}[t]{l} query type: general
    \end{tabular} \\ 
\end{tabular} & 20.91\% \\
\begin{tabular}{ll}
    & \begin{tabular}[t]{l} query type: specific
    \end{tabular} \\ 
\end{tabular} & 52.37\% \\
\begin{tabular}{ll}
    & \begin{tabular}[t]{l} query type: yes/no
    \end{tabular} \\ 
\end{tabular} & 26.72\% \\
\cmidrule{1-2}
\begin{tabular}{ll}
    & \begin{tabular}[t]{l} context-dependent
    \end{tabular} \\ 
\end{tabular} & 13.17\% \\
\begin{tabular}{ll}
 & \begin{tabular}[t]{l} unanswerable
    \end{tabular} \\ 
\end{tabular} & 30.62\% \\
\midrule
avg. query-response pairs per dialog & 9.63 \\
avg. query length (\# words) & 15.22 \\
avg. response length (\# words) & 40.80 \\
\bottomrule
\end{tabular}
\caption{Overall statistics of the MISeD dataset. Queries are classified as either `general', `specific', or `yes/no'. Additionally, queries may be tagged as `context-dependent' and `unanswerable'. 
Table \ref{table:statistics_by_split} provides statistics by dataset split. Figure \ref{fig:question_words} shows an additional breakdown by question words.
}
\label{MISeD-dataset-statistis}
\end{table} 

\paragraph{Splits}
We followed the dataset splits from QMSum, with train, validation and test splits in an approximate ratio of 70:15:15. For each meeting, we aimed to generate two dialogs, each containing ten query-response pairs. Some pairs and dialogs were later filtered during the annotation process.

\paragraph{Context-dependency}
Through manual analysis of a sample of 100 queries annotated as `context-dependent', we observed 3 primary characteristics defining context-dependent queries:
(1) Detail Seeking: 75 queries sought specific details or clarification regarding a prior response (e.g., "\textit{Besides the ease of use, were there any other advantages mentioned for having an LCD screen?}");
(2) Topic Shifting: 17 queries aimed to change topics or explore other aspects of the meeting not previously discussed ("\textit{What other topics did the participants discuss?}" -- response must consider the dialog history to avoid repeating previous topics);
(3) Anaphoric Reference: 13 queries utilized pronouns, necessitating interpretation through preceding dialogue turns ("\textit{What were her recommendations to address this?}"); 
These characteristics can co-occur within a single query ("\textit{What were her other ideas for the input mechanism for the remote?}").

\paragraph{Attribution}
Nearly all (99\%) MISeD responses for answerable queries are supported by transcript attribution, with a median of two attributing transcript spans per response. Attributing spans are relatively long and scattered, with a median length of 96 words, and a median distance of 350 words between subsequent spans (Figure \ref{fig:trancript_and_attributions}).

\paragraph{Process}
Following the process in Section \ref{data-creation-methodology}, we generated 443 dialogs comprising 4430 query-response turns. During validation, annotators eliminated 6\% of the queries, and corrected 11\% of the remaining responses. The average annotation time was 105 minutes per dialog.

\subsection{Wizard-of-Oz Dataset}\label{wizard-of-oz}
To test the value of our semi-automatic MISeD data compared to fully-manual data, we also collected a smaller set of dialogs using the Wizard-of-Oz (WOZ) methodology. As typical in similar WOZ processes (\citealp{quac:2018}, \citealp{topiocqa:2022}), the `user' annotator received a short meeting description, simulating their prior knowledge of the meeting context. They were then instructed to ask free questions to understand aspects of choice about the meeting content.
The `agent' annotator received the full meeting transcript and was tasked with providing free-form answers to the user queries, with supporting attributions. 

The collected WOZ data comprises 70 dialogs based on meetings from the test split, with a total of 700 query-response pairs. This data is used in subsequent evaluations (Section \ref{models-and-results}) for a comparison of model performance.
Table \ref{table:woz_vs_mised} in the Appendix presents a comparison between the WOZ data and the MISeD data, with respect to response vocabulary and length.

WOZ dialog annotation took 161 minutes on average, making it 1.5 times more time-consuming than MISeD.
While in this work we show the feasibility of automating the generation of queries and responses, followed by annotator reviewing, we expect that higher speed-up ratios would be achieved once attribution automation of a reasonable quality is obtained in future research.

\subsection{MISeD Response Quality Assessment} \label{quality-assessment}
MISeD dialogs were generated by an LLM and then validated and corrected by humans (\S \ref{data-creation-methodology}). To assess the quality of the agent responses created through this semi-automatic process, we aimed to compare them with responses created fully-manually.
To that end, we collected human responses to a randomly selected sample of 100 MISeD queries. Annotators were provided with the meeting transcript and the dialog history up to the selected query and were tasked with answering the query.

Next, we provided a new group of annotators with the manually-generated responses, alongside the original MISeD response for each query (in randomized order between the two), and the corresponding query, dialog history, and meeting transcript. Annotators assessed which response was overall superior, considering correctness, grounding, and clarity, using a scale of `Equally Good', `Slightly Better', or `Substantially Better'. Table \ref{MISeD-response-assessment} presents the results.

Overall, MISeD responses were ranked as better in 55 pairs, compared to 43 pairs for fully-manual responses. A statistical sign test yields $P(x\ge55)=0.13$ ($H_0: p = 0.5$; $n=98$). 
These results suggest that the quality of MISeD responses is at least as good as fully-manual responses, and possibly somewhat better, though a larger sample would be needed to establish statistical significance.

\begin{table}
\centering
\small
\begin{tabular}{l>{\centering\arraybackslash}p{1.4cm}>{\centering\arraybackslash}p{0.5cm}}
\textbf{Response evaluation categories} & \textbf{Count} & \textbf{Sum} \\
\toprule 
MISeD `substantially better' & 29 & 55\\
MISeD `slightly better' & 26 & \\ 
\midrule
`equally good' & 2 & \\ 
\midrule
Human-only `slightly better' & 21 & 43\\ 
Human-only `substantially better' & 22 & \\
\bottomrule 
\end{tabular}
\caption{Results of MISeD answer quality assessment. Annotators compared 100 pairs of MISeD and fully-manual responses to the same query, determining the better response in each pair.}
\label{MISeD-response-assessment}
\end{table}
\section{Evaluation Methodologies}\label{sectioneval}
We evaluate the agent models along two dimensions: the quality of the generated responses (\S\ref{section_eval_response}), and the accuracy of the provided attributions (\S\ref{section_eval_attribution}), through both automatic and human evaluations.

\subsection{Response Quality Evaluation} \label{section_eval_response}
Agent responses are evaluated against the gold responses in the test dataset. Modern LLMs produce outputs that very often exhibit high readability and consistency, therefore we chose to focus our evaluation on the content.
\subsubsection{Human evaluation}
We conduct human evaluation on a random subset of 100 queries. The annotator is presented with the current user query, the gold response and the model responses, as well as the meeting transcript and the dialog history.
To quantify the content overlap between the model response and the gold response, annotators provide scores for recall (how much of the gold response is covered by the model response), and precision (how much of the model response is covered by the gold response). Here covered refers to content overlap between the gold and model responses, in a semantic sense. Both scores are on a Likert scale of 1 to 4, corresponding to `Not Covered', `Slightly Covered', `Mostly Covered' and `Fully Covered'.

\subsubsection{Automatic evaluation}
To automatically score model responses we use the standard Rouge-1, Rouge-2 and Rouge-L scores\footnote{https://github.com/google-research/google-research/tree/master/rouge} \citep{lin2004rouge}, for lexical overlap.
To capture semantic overlap, we report BLEURT scores\footnote{https://github.com/google-research/bleurt} \citep{sellam2020bleurt}, a learned evaluation metric based on BERT \citep{devlin-etal-2019-bert}, trained to model human judgments for reference-based text generation evaluation.

\subsection{Attribution Quality Evaluation}\label{section_eval_attribution}
To evaluate model attribution, we adopt the AIS protocol \citep{rashkin2023measuring} as extended by \citep{liu2023evaluating} and \citep{gao2023enabling}. In this approach, attribution evaluation is modeled as a Textual Entailment (Natural Language Inference) task where the generated response text should be entailed by its attributions.

\subsubsection{Human evaluation}
We sample 100 queries from each test set for human evaluation, after filtering out responses for which it was judged (manually) that attribution is not needed (20\% of all responses).

For each model response, annotators assess recall and precision of attributions.
For recall, we break the response into sentences asking for each sentence whether it is fully supported by the set of attributions (score of 1) or not (score of 0), reporting micro-average over all sentences. 
For precision, we ask for each attribution span whether it entails some piece of information in the model response (“partially supports” the response, score of 1), or not (score of 0), reporting micro-average over all attributions.

\subsubsection{Automatic evaluation}\label{section_eval_attribution_auto}
For automatic attribution evaluation we use the NLI model TrueTeacher \citep{gekhman2023trueteacher} to approximate human entailment judgement, automatically computing recall (the proportion of response sentences fully entailed by the attributions) and precision (the proportion of attributions which contribute to the entailment of the response). Appendix \ref{appendix_attribution_eval} details the method implementation and discusses its limitations.

\section{Baseline Models and Results} \label{models-and-results}

\begin{table*} \small
\centering
\begin{tabularx}{0.95\textwidth}{|clccXcccc|}
\hline 
& & \multicolumn{2}{c}{Response - Human scores} & & \multicolumn{4}{c|}{Response - Automatic scores} \\
\cline{3-4}\cline{6-9}
& Model & recall & precision & & Rouge-1 & Rouge-2 & Rouge-L & BLEURT \\
\hline 
\multirow{4}{2.3em}{MISeD} & LongT5 Finetuned & 2.38 & 2.52 & & 44.59 & 27.30 & 37.62 & 0.47\\ 
& Gemini Pro & 2.79 & 2.71 & & 44.64 & 27.49 & 37.35 & 0.48 \\ 
& Gemini Ultra & 2.63 & \textbf{2.87} & & 44.20 & 26.58 & 37.39 & 0.47 \\
& Gemini Pro Finetuned & \textbf{2.96} & 2.86 & & \textbf{51.02} & \textbf{33.38} & \textbf{43.03} & \textbf{0.52}\\ 
\hline 
\multirow{4}{2.3em}{WOZ} & LongT5 Finetuned & 1.79 & 1.80 & & 26.84 & 8.95 & 21.22 & 0.37 \\ 
& Gemini Pro & 2.10 & 2.08 & & 27.82 & 10.08 & 22.03 & 0.38 \\ 
& Gemini Ultra & 1.98 & \textbf{2.14} & & 28.57 & 10.99 & 23.64 & 0.38 \\
& Gemini Pro Finetuned & \textbf{2.21} & 2.13 & & \textbf{30.31} & \textbf{11.39} & \textbf{24.26} & \textbf{0.40} \\ 
\hline 
\end{tabularx}
\caption{Response evaluation: average scores for the MISeD and WOZ test sets.}
\label{result_response}
\end{table*} 

\begin{table*} \small
\centering
\begin{tabularx}{0.85\textwidth}{@{\extracolsep{2.5pt}}|clcccXccc|@{}}
\hline 
& & \multicolumn{3}{c}{Attribution - Human scores} & & \multicolumn{3}{c|}{Attribution - Automatic scores} \\
\cline{3-5}\cline{7-9}
& Model & recall & precision & F1 & & recall & precision & F1\\
\hline 
\multirow{4}{2.3em}{MISeD} & LongT5 Finetuned & \textbf{0.76} & 0.58 & 0.66 & & 0.44 & 0.19 & 0.27 \\
& Gemini Pro & 0.20 & 0.69 & 0.31 & & 0.26 & \textbf{0.37} & 0.31 \\
& Gemini Ultra & 0.19 & \textbf{0.96} & 0.32 & & 0.12 & 0.34 & 0.18 \\
& Gemini Pro Finetuned & 0.68 & 0.69 & \textbf{0.68} & & \textbf{0.50} & 0.27 & \textbf{0.35} \\
\hline 
\multirow{4}{2.3em}{WOZ} & LongT5 Finetuned & \textbf{0.77} & 0.54 & 0.63 & & \textbf{0.51} & 0.17 & 0.26 \\
& Gemini Pro & 0.43 & \textbf{0.85} & 0.57 & & 0.31 & 0.37 & \textbf{0.34} \\
& Gemini Ultra & 0.19 & 0.71 & 0.30 & & 0.13 & \textbf{0.40} & 0.20 \\
& Gemini Pro Finetuned & 0.74 & 0.67 & \textbf{0.70} & & 0.50 & 0.25 & 0.33 \\
\hline 
\end{tabularx}
\caption{Attribution evaluation: average scores for the MISeD and WOZ test sets.}
\label{result_attribution}
\end{table*} 

\subsection{Models}
As defined in Section \ref{sectiontask}, our task input contains the full meeting transcript and the dialog history, ending with the current user query. The output is a concatenation of the response and the set of attributions (indices of supporting segments within the transcript).
In the rare occurrences where the context exceeds the model input size capacity, the beginning of the transcript is truncated.
Results are compared for the following three model types.

\paragraph{Finetuned Encoder-Decoder} LongT5 \citep{guo2021longt5} is a T5 \citep{t5} variant that uses transient global attention (windowing token averaging) to handle longer input contexts efficiently. We finetuned the open-source\footnote{https://github.com/google-research/longt5} LongT5 XL (3 billion parameters) on the MISeD training set, using a context length of 16 thousand tokens.

\paragraph{LLMs prompting} we use the Gemini Pro model and the much larger Gemini Ultra model \citep{team2023gemini},\footnote{https://gemini.google.com/app} without any additional tuning. Our prompt contains the transcript, an instruction, and the dialog ending with the user query (see Table \ref{table_inference_prompt} in the Appendix). The models 28 thousand tokens context limit makes it unfeasible to include few-shots examples (which would require providing additional transcripts with corresponding dialogs).

\paragraph{Finetuned LLM} we finetune the Gemini Pro model\footnote{https://ai.google.dev/docs/model\_tuning\_guidance} on the MISeD training set, using the same prompt and context length as for the prompting approach. The target format is given in Table \ref{table_target}.

\begin{table*} \small
\centering
\setlength{\tabcolsep}{3pt}
\begin{tabular}{|lccccccccc|}
\hline 
& \multicolumn{2}{c}{Response - Human scores} & \multicolumn{4}{c}{Response - Automatic scores} & \multicolumn{3}{c|}{Attribution - Automatic scores}\\
\cline{2-3}\cline{4-7}\cline{8-10}
Training Set & recall & precision & Rouge-1 & Rouge-2 & Rouge-L & BLEURT & recall & precision & F1\\
\hline 
MISeD (n=2922) & 1.91 & 1.97 &  29.73 & 10.90 & 23.64 & 0.40 & 0.28 & 0.25 & 0.26 \\
MISeD (n=500) & \textbf{1.93} & \textbf{1.99} & 30.52 & 11.10 & 24.10 & 0.40 
& 0.26 & 0.16 & 0.20
\\
WOZ (n=500) & 1.91 & 1.94 & \textbf{31.88} & \textbf{12.64} & \textbf{26.17} & \textbf{0.41} &  0.02 & 0.36 & 0.04\\
\hline 
\end{tabular}
\caption{Evaluation using different training sets: the semi-automatic MISeD data and the WOZ manual data. Results on 200 WOZ test examples, using Gemini Pro finetuned.}
\label{result_trainingsets}
\end{table*} 

\subsection{Datasets}
To assess the value of our dataset, in Section \ref{results-dialog} we fintune the agent models on MISeD data, and test it on both MISeD and Wizard-of-Oz (\S\ref{wizard-of-oz}) test sets.
In Section \ref{results-training-comparison} we compare training our best model on the semi-automatic MISeD data to training it on the manually created WOZ data.
Section \ref{results-qmsum} reports results on the QMSum query-based summarization test set.

\subsection{Results for Dialog Data} \label{results-dialog}
\subsubsection{Response quality}
Table \ref{result_response} reports response quality results on the MISeD and WOZ test sets. Automatic evaluation covered the full test sets (628 MISeD queries, 700 WOZ queries), while manual evaluation was performed on a random subset of 100 queries from each.
Our main takeaways are:

(1) The LongT5 model finetuned on MISeD performs similarly to much larger models (about 0.3 points difference in the 4-points human score), highlighting a key advantage of MISeD as it nearly closes the gap between the smaller 3B-parameter model and the much larger Gemini models.

(2) Finetuning Gemini Pro on MISeD significantly improves its performance, surpassing even the much larger Gemini Ultra model, demonstrating MISeD's effectiveness in boosting performance even for large models.

(3) The scores on WOZ test set exhibit the same trends, but are lower for all models. The lower performance for the non-finetuned models might suggest that fully-manual WOZ data is more challenging, being created freely by human annotators, compared to the templated methodology of MISeD. As expected, the finetuned models performed better on the MISeD test set, as it was created by the same protocol as the training data.

\begin{table*} \small
\centering
\begin{tabular}{@{\extracolsep{6pt}}|lcccc|@{}}
\hline 
Model & Rouge-1 & Rouge-2 & Rouge-L & BLEURT \\
\hline
QMSum paper (using retriever) & 32.29 & 8.67 & \textbf{28.17} &   \\
Gemini Ultra & 31.52 & 9.89 & 20.88 & 0.36 \\
\hline
LongT5 Finetuned on QMSum & 35.34 & 12.15 & 23.87 & 0.35 \\
LongT5 Finetuned on MISeD & 29.64 & 8.19 & 20.07 & 0.34 \\
LongT5 Finetuned on both MISeD and QMSum & 35.84 & 12.73 & 24.37 & 0.36 \\
\hline
Gemini Pro & 29.78 & 8.91 & 19.52 & \textbf{0.37} \\
Gemini Pro Finetuned on QMSum & 36.88 & 12.83 & 24.52 & \textbf{0.37} \\
Gemini Pro Finetuned on MISeD & 32.07 & 9.19 & 21.14 & 0.36 \\
Gemini Pro Finetuned on both MISeD and QMSum & \textbf{36.98} & \textbf{13.62} & 25.21 & \textbf{0.37} \\
\hline 
\end{tabular}
\caption{Response evaluation for the QMSum test set.}
\label{eval_response_qmsum}
\end{table*}

\begin{table*} \small
\centering
\begin{tabular}{@{\extracolsep{6pt}}|lcccc|@{}}
\hline 
Model & Rouge-1 & Rouge-2 & Rouge-L & BLEURT \\
\hline 
Gemini Pro & 27.82 & 10.08 & 22.03 & 0.38 \\ 
Gemini Pro Finetuned on MISeD & \textbf{30.31} & \textbf{11.39} & \textbf{24.26} & \textbf{0.40} \\ 
Gemini Pro Finetuned on QMSum & 27.72 & 8.89 & 21.60 & 0.38 \\
\hline 
\end{tabular}
\caption{Average scores for response evaluation on the WOZ test sets.}
\label{eval_response_trainqmsum}
\end{table*} 

\subsubsection{Attribution quality}\label{result_attribution_subsection}
Attribution evaluation results are presented in Table \ref{result_attribution}.
In line with previous work \citep{gao2023enabling}, our results suggest that pre-trained LLMs do not excel at finding attributions over meeting transcripts.\footnote{This is consistent with our findings when developing the MISeD annotation methodology (\S\ref{sec_dataset_annotation}), which led us to leave attribution generation fully manual at this point.}

Observing the more reliable human scores, we see that the zero-shot models suffer from low recall, due to their tendency to provide attribution less frequently. Finetuning with MISeD data improves performance, substantially increasing recall with a small drop in precision, notably increasing F1 scores for both test sets.
The automatic scores show similar though weaker trends, but we regard them as less reliable (see Appendix \ref{appendix_attribution_eval}).

\subsection{Comparison of Training Sets} \label{results-training-comparison}

In this section we compare the performance of models trained on our semi-automatic MISeD data to those trained on the manual WOZ data. We split our WOZ dataset (700 dialog turns) into training and test sets (500:200, respectively). We compare training with (i) 500 WOZ examples (ii) 500 semi-automatic MISeD examples (iii) the full MISeD training set (2922 examples). For response generation, Table \ref{result_trainingsets} suggests that training on 500 semi-automatic MISeD examples yields comparable results to training on the same amount of fully manual examples.
For attribution, the model benefits from a larger number of training examples, suggesting that 500 examples are insufficient for this subtask. Notably, the model trained on WOZ only predicted attributions for 6 out of 200 test examples; thus we didn’t perform a manual attribution evaluation (as in Table \ref{result_attribution}).
We also tried adding the WOZ data to the MiSeD data in finetuning, but, with respect to automatic evaluation, this did not improve notably the response quality while attribution quality was notably deteriorated, hence we did not pursue this data combination further.
To summarize, training on our semi-automatic MISeD data yields response generation quality that is on par with training on the fully manual WOZ data, while also improving attribution, and saving time and effort. 

\subsection{Results for the QMSum Data} \label{results-qmsum}
For comparison with existing results, we report results on the QMSum single-turn query-based summarization test set. We compare the model response quality when finetuned on (i) MISeD training data (ii) QMSum training data (iii) both MISeD and QMSum training data together (attribution evaluation is not included in the original work).

Our results (Table \ref{eval_response_qmsum}) indicate that finetuning with MISeD has additive benefits: when the model learns on both MISeD and QMSum data, it surpasses zero-shot models, as well as models trained on QMSum alone or MISeD alone. Our best model, Gemini Pro finetuned on both MISeD and QMSum data, surpasses the performances reported in the original QMSum study for Rouge-1 and Rouge-2.
As expected, models trained solely on MISeD do not outperform QMSum-trained models on their own test data: MISeD data is out-of-distribution as it deviates significantly from the QMSum schema, designed for fully manual annotation.

Additionally, Table \ref{eval_response_trainqmsum} shows that training on the single-turn QMSum data yields a lower response quality compared to training on MISeD dialog data, while testing on WOZ dialogs. Furthermore, finetuning with QMSum does not improve at all over the non-finetuned model. This demonstrates that non-dialog data composed of standalone question-answer pairs has a limited value when training models for dialog tasks.

\section{Conclusion}
In this paper, we investigate LLM automation to generate source-grounded information-seeking dialog datasets. We introduce a method to partially automate the WOZ process using targeted user and agent prompts, followed by human attribution, verification, and potential editing. We apply this method to create MISeD, the first dataset for information-seeking dialogs over meetings.

Baseline models and experiments demonstrate the value of MISeD: finetuninig with MISeD data brings the same improvement as a comparable amount of the more expensive fully-manual WOZ data. MISeD enables the creation of modestly sized finetuned encoder-decoder models with performance approaching much larger pre-trained LLMs, and further improves the performance of such LLMs through finetuning.
We suggest that our work yields valuable insights about the feasibility of LLM-based data generation in the important and challenging case of source-grounded information-seeking dialogs.

\section{Limitations}
\paragraph{Attribution}
Our work successfully automates query and response generation, but highlights the challenge of attribution in both modeling and automatic evaluation. Future research on improving attribution generation models could enable fuller automation of dialogs generation, leading to an even more efficient process.

\paragraph{Long texts depend on LLM context length}
Our method's reliance on LLM prompting poses challenges for long meeting transcripts that exceed the maximal context length of the model. The same problem occurs in the response generation agent task.
Possible solutions include using LLMs with larger context lengths (which may be computationally expensive) or integrating a retrieval system as a first stage in the response generation, to condense the transcript and provide the most relevant information to the LLM.

\paragraph{Manually crafted prompt templates} 
Our method currently depends on manually crafted prompt templates to represent a broad spectrum of potential queries in the given domain. Research into more flexible and diverse query generation strategies could reduce the effort of applying our method in new domains while increasing the generality of the obtained data.
\section*{Acknowledgements}
We wish to thank David Karam, Michelle Tadmor Ramanovich, Eliya Nachmani, Benny Schlesinger, David Petru and Blaise Aguera y Arcas for their support and feedback on the research. We also would like to express our gratitude to Victor Cărbune, Lucas Werner and Ondrej Skopek for their help with the annotation framework, and to Riteeka Kapila for leading the annotators team.

\bibliography{anthology,custom}

\begin{thebibliography}{43}
\expandafter\ifx\csname natexlab\endcsname\relax\def\natexlab#1{#1}\fi

\bibitem[{Adlakha et~al.(2022)Adlakha, Dhuliawala, Suleman, de~Vries, and
  Reddy}]{topiocqa:2022}
Vaibhav Adlakha, Shehzaad Dhuliawala, Kaheer Suleman, Harm de~Vries, and Siva
  Reddy. 2022.
\newblock \href {http://arxiv.org/abs/2110.00768} {Topiocqa: Open-domain
  conversational question answering with topic switching}.

\bibitem[{Anantha et~al.(2021)Anantha, Vakulenko, Tu, Longpre, Pulman, and
  Chappidi}]{QReCC:2021}
Raviteja Anantha, Svitlana Vakulenko, Zhucheng Tu, Shayne Longpre, Stephen
  Pulman, and Srinivas Chappidi. 2021.
\newblock \href {http://arxiv.org/abs/2010.04898} {Open-domain question
  answering goes conversational via question rewriting}.

\bibitem[{Apel et~al.(2023)Apel, Braude, Kantor, and Kolman}]{apel2023meeqa}
Reut Apel, Tom Braude, Amir Kantor, and Eyal Kolman. 2023.
\newblock \href {http://arxiv.org/abs/2305.08502} {Meeqa: Natural questions in
  meeting transcripts}.

\bibitem[{Bao et~al.(2023)Bao, Wang, Wang, Sun, Li, Mi, and
  Xu}]{synthetic:2023}
Jianzhu Bao, Rui Wang, Yasheng Wang, Aixin Sun, Yitong Li, Fei Mi, and Ruifeng
  Xu. 2023.
\newblock \href {https://doi.org/10.18653/v1/2023.acl-long.608} {A synthetic
  data generation framework for grounded dialogues}.
\newblock In \emph{Proceedings of the 61st Annual Meeting of the Association
  for Computational Linguistics (Volume 1: Long Papers)}, pages 10866--10882,
  Toronto, Canada. Association for Computational Linguistics.

\bibitem[{Budzianowski et~al.(2018)Budzianowski, Wen, Tseng, Casanueva, Ultes,
  Ramadan, and Ga{\v{s}}i{\'c}}]{multiwoz:2018}
Pawe{\l} Budzianowski, Tsung-Hsien Wen, Bo-Hsiang Tseng, I{\~n}igo Casanueva,
  Stefan Ultes, Osman Ramadan, and Milica Ga{\v{s}}i{\'c}. 2018.
\newblock \href {https://doi.org/10.18653/v1/D18-1547} {{M}ulti{WOZ} - a
  large-scale multi-domain {W}izard-of-{O}z dataset for task-oriented dialogue
  modelling}.
\newblock In \emph{Proceedings of the 2018 Conference on Empirical Methods in
  Natural Language Processing}, pages 5016--5026, Brussels, Belgium.
  Association for Computational Linguistics.

\bibitem[{Campos et~al.(2020)Campos, Otegi, Soroa, Deriu, Cieliebak, and
  Agirre}]{doqa:2020}
Jon~Ander Campos, Arantxa Otegi, Aitor Soroa, Jan Deriu, Mark Cieliebak, and
  Eneko Agirre. 2020.
\newblock \href {http://arxiv.org/abs/2005.01328} {Doqa -- accessing
  domain-specific faqs via conversational qa}.

\bibitem[{Carletta et~al.(2005)Carletta, Ashby, Bourban, Flynn, Guillemot,
  Hain, Kadlec, Karaiskos, Kraaij, Kronenthal, Lathoud, Lincoln, Lisowska,
  McCowan, Post, Reidsma, and Wellner}]{ami:2005}
Jean Carletta, Simone Ashby, Sebastien Bourban, Mike Flynn, Ma{\"{e}}l
  Guillemot, Thomas Hain, Jaroslav Kadlec, Vasilis Karaiskos, Wessel Kraaij,
  Melissa Kronenthal, Guillaume Lathoud, Mike Lincoln, Agnes Lisowska, Iain
  McCowan, Wilfried Post, Dennis Reidsma, and Pierre Wellner. 2005.
\newblock \href {https://doi.org/10.1007/11677482\_3} {The {AMI} meeting
  corpus: {A} pre-announcement}.
\newblock In \emph{Machine Learning for Multimodal Interaction, Second
  International Workshop, {MLMI} 2005, Edinburgh, UK, July 11-13, 2005, Revised
  Selected Papers}, volume 3869 of \emph{Lecture Notes in Computer Science},
  pages 28--39. Springer.

\bibitem[{Chen et~al.(2023)Chen, Papangelis, Tao, Kim, Rosenbaum, Liu, Yu, and
  Hakkani-Tur}]{chen2023places}
Maximillian Chen, Alexandros Papangelis, Chenyang Tao, Seokhwan Kim, Andy
  Rosenbaum, Yang Liu, Zhou Yu, and Dilek Hakkani-Tur. 2023.
\newblock Places: Prompting language models for social conversation synthesis.
\newblock In \emph{Findings of the Association for Computational Linguistics:
  EACL 2023}, pages 844--868.

\bibitem[{Chen et~al.(2022)Chen, Papangelis, Tao, Rosenbaum, Kim, Liu, Yu, and
  Hakkani-T{\"u}r}]{Chen2022WeaklySD}
Maximillian Chen, Alexandros Papangelis, Chenyang Tao, Andrew Rosenbaum,
  Seokhwan Kim, Yang Liu, Zhou Yu, and Dilek~Z. Hakkani-T{\"u}r. 2022.
\newblock \href {https://api.semanticscholar.org/CorpusID:253107809} {Weakly
  supervised data augmentation through prompting for dialogue understanding}.
\newblock \emph{ArXiv}, abs/2210.14169.

\bibitem[{Choi et~al.(2018)Choi, He, Iyyer, Yatskar, Yih, Choi, Liang, and
  Zettlemoyer}]{quac:2018}
Eunsol Choi, He~He, Mohit Iyyer, Mark Yatskar, Wen-tau Yih, Yejin Choi, Percy
  Liang, and Luke Zettlemoyer. 2018.
\newblock \href {https://doi.org/10.18653/v1/D18-1241} {{Q}u{AC}: Question
  answering in context}.
\newblock In \emph{Proceedings of the 2018 Conference on Empirical Methods in
  Natural Language Processing}, pages 2174--2184, Brussels, Belgium.
  Association for Computational Linguistics.

\bibitem[{Devlin et~al.(2019)Devlin, Chang, Lee, and
  Toutanova}]{devlin-etal-2019-bert}
Jacob Devlin, Ming-Wei Chang, Kenton Lee, and Kristina Toutanova. 2019.
\newblock \href {https://doi.org/10.18653/v1/N19-1423} {{BERT}: Pre-training of
  deep bidirectional transformers for language understanding}.
\newblock In \emph{Proceedings of the 2019 Conference of the North {A}merican
  Chapter of the Association for Computational Linguistics: Human Language
  Technologies, Volume 1 (Long and Short Papers)}, pages 4171--4186,
  Minneapolis, Minnesota. Association for Computational Linguistics.

\bibitem[{Dinan et~al.(2019)Dinan, Roller, Shuster, Fan, Auli, and
  Weston}]{wizardofwikipedia:2019}
Emily Dinan, Stephen Roller, Kurt Shuster, Angela Fan, Michael Auli, and Jason
  Weston. 2019.
\newblock \href {http://arxiv.org/abs/1811.01241} {Wizard of wikipedia:
  Knowledge-powered conversational agents}.

\bibitem[{Feng et~al.(2020)Feng, Wan, Gunasekara, Patel, Joshi, and
  Lastras}]{doc2dial:2020}
Song Feng, Hui Wan, Chulaka Gunasekara, Siva Patel, Sachindra Joshi, and Luis
  Lastras. 2020.
\newblock \href {https://doi.org/10.18653/v1/2020.emnlp-main.652} {doc2dial: A
  goal-oriented document-grounded dialogue dataset}.
\newblock In \emph{Proceedings of the 2020 Conference on Empirical Methods in
  Natural Language Processing (EMNLP)}, pages 8118--8128, Online. Association
  for Computational Linguistics.

\bibitem[{Gao et~al.(2023)Gao, Yen, Yu, and Chen}]{gao2023enabling}
Tianyu Gao, Howard Yen, Jiatong Yu, and Danqi Chen. 2023.
\newblock Enabling large language models to generate text with citations.
\newblock \emph{arXiv preprint arXiv:2305.14627}.

\bibitem[{Gekhman et~al.(2023)Gekhman, Herzig, Aharoni, Elkind, and
  Szpektor}]{gekhman2023trueteacher}
Zorik Gekhman, Jonathan Herzig, Roee Aharoni, Chen Elkind, and Idan Szpektor.
  2023.
\newblock \href {http://arxiv.org/abs/2305.11171} {Trueteacher: Learning
  factual consistency evaluation with large language models}.

\bibitem[{{Gemini Team Google}(2023)}]{team2023gemini}
{Gemini Team Google}. 2023.
\newblock Gemini: A family of highly capable multimodal models.
\newblock \emph{arXiv preprint arXiv:2312.11805}.

\bibitem[{Gopalakrishnan et~al.(2019)Gopalakrishnan, Hedayatnia, Chen,
  Gottardi, Kwatra, Venkatesh, Gabriel, and
  Hakkani-Tür}]{gopalakrishnan19_interspeech}
Karthik Gopalakrishnan, Behnam Hedayatnia, Qinlang Chen, Anna Gottardi, Sanjeev
  Kwatra, Anu Venkatesh, Raefer Gabriel, and Dilek Hakkani-Tür. 2019.
\newblock \href {https://doi.org/10.21437/Interspeech.2019-3079}
  {{Topical-Chat: Towards Knowledge-Grounded Open-Domain Conversations}}.
\newblock In \emph{Proc. Interspeech 2019}, pages 1891--1895.

\bibitem[{Guo et~al.(2021)Guo, Ainslie, Uthus, Ontanon, Ni, Sung, and
  Yang}]{guo2021longt5}
Mandy Guo, Joshua Ainslie, David Uthus, Santiago Ontanon, Jianmo Ni, Yun-Hsuan
  Sung, and Yinfei Yang. 2021.
\newblock \href {http://arxiv.org/abs/2112.07916} {Longt5: Efficient
  text-to-text transformer for long sequences}.

\bibitem[{Honovich et~al.(2022)Honovich, Aharoni, Herzig, Taitelbaum,
  Kukliansy, Cohen, Scialom, Szpektor, Hassidim, and
  Matias}]{honovich-etal-2022-true-evaluating}
Or~Honovich, Roee Aharoni, Jonathan Herzig, Hagai Taitelbaum, Doron Kukliansy,
  Vered Cohen, Thomas Scialom, Idan Szpektor, Avinatan Hassidim, and Yossi
  Matias. 2022.
\newblock \href {https://doi.org/10.18653/v1/2022.naacl-main.287} {{TRUE}:
  Re-evaluating factual consistency evaluation}.
\newblock In \emph{Proceedings of the 2022 Conference of the North American
  Chapter of the Association for Computational Linguistics: Human Language
  Technologies}, pages 3905--3920, Seattle, United States. Association for
  Computational Linguistics.

\bibitem[{Hu et~al.(2023)Hu, Ganter, Deilamsalehy, Dernoncourt, Foroosh, and
  Liu}]{meetingbank:2023}
Yebowen Hu, Timothy Ganter, Hanieh Deilamsalehy, Franck Dernoncourt, Hassan
  Foroosh, and Fei Liu. 2023.
\newblock \href {https://doi.org/10.18653/V1/2023.ACL-LONG.906} {Meetingbank:
  {A} benchmark dataset for meeting summarization}.
\newblock In \emph{Proceedings of the 61st Annual Meeting of the Association
  for Computational Linguistics (Volume 1: Long Papers), {ACL} 2023, Toronto,
  Canada, July 9-14, 2023}, pages 16409--16423. Association for Computational
  Linguistics.

\bibitem[{Janin et~al.(2003)Janin, Baron, Edwards, Ellis, Gelbart, Morgan,
  Peskin, Pfau, Shriberg, Stolcke, and Wooters}]{icsi:2003}
Adam Janin, Don Baron, Jane Edwards, Dan Ellis, David Gelbart, Nelson Morgan,
  Barbara Peskin, Thilo Pfau, Elizabeth Shriberg, Andreas Stolcke, and Chuck
  Wooters. 2003.
\newblock \href {https://doi.org/10.1109/ICASSP.2003.1198793} {The {ICSI}
  meeting corpus}.
\newblock In \emph{2003 {IEEE} International Conference on Acoustics, Speech,
  and Signal Processing, {ICASSP} '03, Hong Kong, April 6-10, 2003}, pages
  364--367. {IEEE}.

\bibitem[{Kelley(1984)}]{woz:1984}
J.~F. Kelley. 1984.
\newblock \href {https://doi.org/10.1145/357417.357420} {An iterative design
  methodology for user-friendly natural language office information
  applications}.
\newblock \emph{ACM Trans. Inf. Syst.}, 2(1):26–41.

\bibitem[{Kim et~al.(2023)Kim, Cho, and Na}]{kim-etal-2023-explainmeetsum}
Hyun Kim, Minsoo Cho, and Seung-Hoon Na. 2023.
\newblock \href {https://doi.org/10.18653/v1/2023.acl-long.731}
  {{E}xplain{M}eet{S}um: A dataset for explainable meeting summarization
  aligned with human intent}.
\newblock In \emph{Proceedings of the 61st Annual Meeting of the Association
  for Computational Linguistics (Volume 1: Long Papers)}, pages 13079--13098,
  Toronto, Canada. Association for Computational Linguistics.

\bibitem[{Li et~al.(2023)Li, Yin, Yang, Jiang, Li, Cheng, Shang, Jiang, Liu,
  and Yang}]{newsdialogues:2023}
Siheng Li, Yichun Yin, Cheng Yang, Wangjie Jiang, Yiwei Li, Zesen Cheng, Lifeng
  Shang, Xin Jiang, Qun Liu, and Yujiu Yang. 2023.
\newblock \href {https://doi.org/10.18653/v1/2023.findings-acl.224}
  {Newsdialogues: Towards proactive news grounded conversation}.
\newblock In \emph{Findings of the Association for Computational Linguistics:
  ACL 2023}. Association for Computational Linguistics.

\bibitem[{Li et~al.(2022)Li, Chen, Li, Wang, Qian, and
  Yan}]{li2022controllable}
Zekun Li, Wenhu Chen, Shiyang Li, Hong Wang, Jing Qian, and Xifeng Yan. 2022.
\newblock Controllable dialogue simulation with in-context learning.
\newblock In \emph{Findings of the Association for Computational Linguistics:
  EMNLP 2022}, pages 4330--4347.

\bibitem[{Lin(2004)}]{lin2004rouge}
Chin-Yew Lin. 2004.
\newblock Rouge: A package for automatic evaluation of summaries.
\newblock In \emph{Text summarization branches out}, pages 74--81.

\bibitem[{Lin et~al.(2022)Lin, Papangelis, Kim, and
  Hakkani-Tur}]{lin-etal-2022-knowledge}
Yen~Ting Lin, Alexandros Papangelis, Seokhwan Kim, and Dilek Hakkani-Tur. 2022.
\newblock \href {https://aclanthology.org/2022.sigdial-1.3} {Knowledge-grounded
  conversational data augmentation with generative conversational networks}.
\newblock In \emph{Proceedings of the 23rd Annual Meeting of the Special
  Interest Group on Discourse and Dialogue}, pages 26--38, Edinburgh, UK.
  Association for Computational Linguistics.

\bibitem[{Liu et~al.(2023)Liu, Zhang, and Liang}]{liu2023evaluating}
Nelson~F Liu, Tianyi Zhang, and Percy Liang. 2023.
\newblock Evaluating verifiability in generative search engines.
\newblock \emph{arXiv preprint arXiv:2304.09848}.

\bibitem[{Mehri et~al.(2022)Mehri, Altun, and Eskenazi}]{mehri2022lad}
Shikib Mehri, Yasemin Altun, and Maxine Eskenazi. 2022.
\newblock Lad: Language models as data for zero-shot dialog.
\newblock In \emph{Proceedings of the 23rd Annual Meeting of the Special
  Interest Group on Discourse and Dialogue}, pages 595--604.

\bibitem[{Nakamura et~al.(2022)Nakamura, Levy, Tuan, Chen, and
  Wang}]{hybridialogue:2022}
Kai Nakamura, Sharon Levy, Yi-Lin Tuan, Wenhu Chen, and William~Yang Wang.
  2022.
\newblock \href {https://doi.org/10.18653/v1/2022.findings-acl.41}
  {{H}ybri{D}ialogue: An information-seeking dialogue dataset grounded on
  tabular and textual data}.
\newblock In \emph{Findings of the Association for Computational Linguistics:
  ACL 2022}, pages 481--492, Dublin, Ireland. Association for Computational
  Linguistics.

\bibitem[{Nedoluzhko et~al.(2022)Nedoluzhko, Singh, Hled{\'\i}kov{\'a}, Ghosal,
  and Bojar}]{nedoluzhko-etal-2022-elitr}
Anna Nedoluzhko, Muskaan Singh, Marie Hled{\'\i}kov{\'a}, Tirthankar Ghosal,
  and Ond{\v{r}}ej Bojar. 2022.
\newblock \href {https://aclanthology.org/2022.lrec-1.340} {{ELITR} minuting
  corpus: A novel dataset for automatic minuting from multi-party meetings in
  {E}nglish and {C}zech}.
\newblock In \emph{Proceedings of the Thirteenth Language Resources and
  Evaluation Conference}, pages 3174--3182, Marseille, France. European
  Language Resources Association.

\bibitem[{Prasad et~al.(2023)Prasad, Bui, Yoon, Deilamsalehy, Dernoncourt, and
  Bansal}]{meetingqa:2023}
Archiki Prasad, Trung Bui, Seunghyun Yoon, Hanieh Deilamsalehy, Franck
  Dernoncourt, and Mohit Bansal. 2023.
\newblock \href {https://doi.org/10.18653/v1/2023.acl-long.837} {{M}eeting{QA}:
  Extractive question-answering on meeting transcripts}.
\newblock In \emph{Proceedings of the 61st Annual Meeting of the Association
  for Computational Linguistics (Volume 1: Long Papers)}, pages 15000--15025,
  Toronto, Canada. Association for Computational Linguistics.

\bibitem[{Raffel et~al.(2020)Raffel, Shazeer, Roberts, Lee, Narang, Matena,
  Zhou, Li, and Liu}]{t5}
Colin Raffel, Noam Shazeer, Adam Roberts, Katherine Lee, Sharan Narang, Michael
  Matena, Yanqi Zhou, Wei Li, and Peter~J. Liu. 2020.
\newblock Exploring the limits of transfer learning with a unified text-to-text
  transformer.
\newblock \emph{J. Mach. Learn. Res.}, 21(1).

\bibitem[{Rashkin et~al.(2023)Rashkin, Nikolaev, Lamm, Aroyo, Collins, Das,
  Petrov, Tomar, Turc, and Reitter}]{rashkin2023measuring}
Hannah Rashkin, Vitaly Nikolaev, Matthew Lamm, Lora Aroyo, Michael Collins,
  Dipanjan Das, Slav Petrov, Gaurav~Singh Tomar, Iulia Turc, and David Reitter.
  2023.
\newblock Measuring attribution in natural language generation models.
\newblock \emph{Computational Linguistics}, pages 1--64.

\bibitem[{Reddy et~al.(2019)Reddy, Chen, and Manning}]{coqa:2019}
Siva Reddy, Danqi Chen, and Christopher~D. Manning. 2019.
\newblock \href {https://doi.org/10.1162/tacl_a_00266} {{C}o{QA}: A
  conversational question answering challenge}.
\newblock \emph{Transactions of the Association for Computational Linguistics},
  7:249--266.

\bibitem[{Saeidi et~al.(2018)Saeidi, Bartolo, Lewis, Singh, Rockt{\"a}schel,
  Sheldon, Bouchard, and Riedel}]{sharc:2018}
Marzieh Saeidi, Max Bartolo, Patrick Lewis, Sameer Singh, Tim Rockt{\"a}schel,
  Mike Sheldon, Guillaume Bouchard, and Sebastian Riedel. 2018.
\newblock \href {https://doi.org/10.18653/v1/D18-1233} {Interpretation of
  natural language rules in conversational machine reading}.
\newblock In \emph{Proceedings of the 2018 Conference on Empirical Methods in
  Natural Language Processing}, pages 2087--2097, Brussels, Belgium.
  Association for Computational Linguistics.

\bibitem[{Sellam et~al.(2020)Sellam, Das, and Parikh}]{sellam2020bleurt}
Thibault Sellam, Dipanjan Das, and Ankur~P Parikh. 2020.
\newblock Bleurt: Learning robust metrics for text generation.
\newblock \emph{arXiv preprint arXiv:2004.04696}.

\bibitem[{Sun et~al.(2021)Sun, Moon, Crook, Roller, Silvert, Liu, Wang, Liu,
  Cho, and Cardie}]{sun2021adding}
Kai Sun, Seungwhan Moon, Paul~A Crook, Stephen Roller, Becka Silvert, Bing Liu,
  Zhiguang Wang, Honglei Liu, Eunjoon Cho, and Claire Cardie. 2021.
\newblock Adding chit-chat to enhance task-oriented dialogues.
\newblock In \emph{Proceedings of the 2021 Conference of the North American
  Chapter of the Association for Computational Linguistics: Human Language
  Technologies}, pages 1570--1583.

\bibitem[{Wu et~al.(2022{\natexlab{a}})Wu, Madotto, Liu, Fung, and
  Xiong}]{qaconv:2022}
Chien-Sheng Wu, Andrea Madotto, Wenhao Liu, Pascale Fung, and Caiming Xiong.
  2022{\natexlab{a}}.
\newblock \href {https://doi.org/10.18653/v1/2022.acl-long.370} {{QAC}onv:
  Question answering on informative conversations}.
\newblock In \emph{Proceedings of the 60th Annual Meeting of the Association
  for Computational Linguistics (Volume 1: Long Papers)}, pages 5389--5411,
  Dublin, Ireland. Association for Computational Linguistics.

\bibitem[{Wu et~al.(2022{\natexlab{b}})Wu, Feng, Chen, Joshi, Lastras, and
  Yu}]{wu-etal-2022-dg2}
Qingyang Wu, Song Feng, Derek Chen, Sachindra Joshi, Luis Lastras, and Zhou Yu.
  2022{\natexlab{b}}.
\newblock \href {https://aclanthology.org/2022.sigdial-1.21} {{DG}2: Data
  augmentation through document grounded dialogue generation}.
\newblock In \emph{Proceedings of the 23rd Annual Meeting of the Special
  Interest Group on Discourse and Dialogue}, pages 204--216, Edinburgh, UK.
  Association for Computational Linguistics.

\bibitem[{Zhang et~al.(2023)Zhang, Deng, Liu, Yu, Chen, Wang, Yan, Liu, Ren,
  and Zhao}]{zhang2023mug}
Qinglin Zhang, Chong Deng, Jiaqing Liu, Hai Yu, Qian Chen, Wen Wang, Zhijie
  Yan, Jinglin Liu, Yi~Ren, and Zhou Zhao. 2023.
\newblock \href {http://arxiv.org/abs/2303.13939} {Mug: A general meeting
  understanding and generation benchmark}.

\bibitem[{Zheng et~al.(2022)Zheng, Sabour, Wen, Zhang, and
  Huang}]{Zheng2022AugESCDA}
Chujie Zheng, Sahand Sabour, Jiaxin Wen, Zheng Zhang, and Minlie Huang. 2022.
\newblock \href {https://api.semanticscholar.org/CorpusID:258588110} {Augesc:
  Dialogue augmentation with large language models for emotional support
  conversation}.
\newblock In \emph{Annual Meeting of the Association for Computational
  Linguistics}.

\bibitem[{Zhong et~al.(2021)Zhong, Yin, Yu, Zaidi, Mutuma, Jha, Awadallah,
  Celikyilmaz, Liu, Qiu, and Radev}]{qmsum:21}
Ming Zhong, Da~Yin, Tao Yu, Ahmad Zaidi, Mutethia Mutuma, Rahul Jha,
  Ahmed~Hassan Awadallah, Asli Celikyilmaz, Yang Liu, Xipeng Qiu, and
  Dragomir~R. Radev. 2021.
\newblock \href {http://arxiv.org/abs/2104.05938} {Qmsum: {A} new benchmark for
  query-based multi-domain meeting summarization}.
\newblock \emph{CoRR}, abs/2104.05938.

\end{thebibliography}
\bibliographystyle{acl_natbib}

\appendix

\section{Prompts additional information} \label{appendix-a}

Our prompt structure consists of generic instructions that provide guidance to the LLM, helping it formulate user queries based on the provided meeting transcript and the conversational history (Table \ref{query_prompt}), followed by the user query template to use in the specific turn (Table \ref{full_query_templates_list}).

Adapted from QMSum, we include both \textit{General} queries for overall meeting themes and key takeaways, and \textit{Specific} queries which focus on particular topics or individuals discussed.
We expanded the QMSum schema based on a preliminary user study, which allowed volunteers to upload their meeting recordings and inquire a prototype agent model about them. Analyzing this interaction data, we added an \textit{Unanswerable} and \textit{Context-dependent} query templates.

Unlike human annotators who can create tailored queries for a given transcript, our model is guided to always generate queries from any prompt. Thus, we adapt the schema towards more generic prompts to ensure the model always produces relevant queries.

For example, QMSum authors suggest the following query in their schema: \textit{"Why did A agree / disagree with B when discussing X?"}. This assumes a disagreement between speaker A and B around some topic X. This assumption will not necessarily be true for any given meeting. Therefore we changed the prompt to be: 
\textit{"The question should be a rephrase of asking if anyone disagreed with <Speaker> about <Topic>"}.

\section{Annotation details} \label{appendix-annotations}
MISeD was annotated by 3 professional annotators, fluent in English, employed in a commercial organization which provides data annotation services. The training process included detailed instructions and a pilot session, over held-out data which included 525 dialog turns, following which we clarified misunderstandings raised in the process. MISeD instances were divided between the 3 annotators, so that each instance was annotated by a single annotator.




\section{Automatic Evaluation of Attribution Implementation} \label{appendix_attribution_eval}
Automatic evaluation of attribution is a recent research area with no established methods in the literature.
We follow the methodology proposed recently by \citep{gao2023enabling} which applies an NLI model to approximate human entailment judgements, automatically computing recall (the proportion of response sentences fully entailed by the attributions) and precision (the proportion of attributions which contribute to the entailment of the response). 
While they utilized the True NLI model \citep{honovich-etal-2022-true-evaluating}, we use the newer NLI model TrueTeacher \citep{gekhman2023trueteacher}.

We compute precision and recall scores for attribution in the following way. 

\textbf{Recall}: recall measures to what extent the generated text is entailed by the attribution. For each sentence $s$ in the response $S$, and given an attribution set $A$ (the concatenation of all transcript spans $a$ provided by the model as attribution), we compute $NLI(A,s)$ as the TrueTeacher-computed score of $A$ entailing $s$ (either 1 or 0). We then average this score over all sentences (micro-average).

\textbf{Precision}: precision measures whether the attribution only includes citations needed to entail the response. We compute citation precision for each citation $a$ of the attribution set $A$. The precision of $a$ is either 0 or 1. We first determine if $a$ is relevant or irrelevant with respect to every sentence $s$ in the response $S$: the citation $a$ should be considered irrelevant wrt $s$ if it does not entail $s$ by itself ($NLI(a,s)=0$) and the overall NLI score of $A$ vs $s$ does not change removing $a$ from $A$ ($NLI(A\setminus\{a\},s)=NLI(A,s)$). $a$ has a precision of 1 if $a$ is relevant for at least a sentence $s$ which has recall=1 ($NLI(A,s)=1$).
We then average the precision of all attributions (micro-average).

As pointed out in \citep{gao2023enabling}, a limitation of this citation precision evaluation is that it cannot detect a citation that partially supports the statement, therefore resulting in lower scores compared to human evaluation. We see this effect in Table \ref{result_attribution} which shows a gap between human-computed and automatically-computed attribution precision. 

As also noted in \citep{gao2023enabling}, not all model responses require attributions. While in their settings responses not requiring attribution were rare and ignored, in the meeting domain we find this case to be more prevalent, particularly when the response indicates that no answer can be found in the transcript (see Section \ref{data-creation-methodology}), an example being "\textit{The meeting participants did not discuss specific ideas for the remote control's shape and size .}".

In the human evaluation we filter out, for each model, from attribution evaluation those responses for which it was judged (manually) that attribution is not needed (20\% of all responses).
Unlike the manual evaluation process, there is no available automatic method to filter out responses that do not need attribution, which remains a challenge for future research. Because the automatic method includes also responses which do not require attribution, we can see a gap between the automatic and manual scores.

Finally we note that while the attribution format is consistent for fine-tuned models, zero-shot models present some inconsistency in how they list attribution and thus they require more complex parsing logic.

\begin{table*}[ht]
\centering
\begin{tabular}{p{6in}}
\hline
\textbf{User query full prompt} \\
\small \textit{<Meeting transcript>} \\
\small Instruction: Generate a multi turn dialog between a user and a chatbot about the provided meeting. Every turn the user asks a question about the meeting and the chatbot respond with an answer based on the meeting \\
\small \textit{<Dialog history>} \\
\small \textit{<Query template>} \\
\hline
\end{tabular}
\caption{Our query prompt includes generic instructions that guide the LLM in formulating user queries, given the meeting transcript and dialog history. This is followed by a randomly chosen query template, from a pool of templates that are based on the QMSum schema and our edits and additions.}
\label{query_prompt}
\end{table*}

\begin{table*}[ht]
\centering
\begin{tabular}{p{6in}}
\hline
\small \textbf{General query templates} \\
\small 1. The question should be a rephrase of asking for a summary of the meeting.\\
\small 2. The question should be a rephrase of asking for a summary of the things <Speaker> said in the meeting.\\
\small 3. The question should be a rephrase of asking what was the  conclusion of the meeting.\\
\small 4. The question should be a rephrase of asking what was the purpose of the meeting.\\
\small 5. The question should be a rephrase of asking what were the action items of the meeting.\\
\small 6. The question should be a rephrase of asking to identify questions raised during the meeting that were left unresolved. \\ 
\hline
\small \textbf{Specific query templates} \\
\small 1. The question should be a rephrase of asking for a summary of <Topic>. <Topic> should be a topic that was discussed within the meeting.\\
\small 2. The question should be a rephrase of asking why <Decision> was made.\\
\small 3. The question should be a rephrase of asking what did <Speaker> say regarding <Topic> in the meeting.\\
\small 4. The question should be a rephrase of asking what was the advantage of <Solution>.\\
\small 5. The question should be a rephrase of asking why <Speaker> held an <Opinion>.\\
\small 6. The question should be a rephrase of asking what was decided regarding <Topic>.\\
\small 7. The question should should be a rephrase of asking if anyone disagreed with <Speaker> about <Topic>.\\
\small 8. The question should be a rephrase of asking what did <Speaker> recommend to do when discussing <Topic>.\\
\small 9. The question should be a Yes/No question where the answer is "Yes" based on the meeting.\\
\small 10. The question should be a Yes/No question where the answer is "No" based on the meeting.\\
\small 11. The question should be a Yes/No question where the answer can not be found within the meeting. \\
\hline
\small \textbf{Unanswerable query templates} \\
\small 1. The question should be a rephrase of asking for a summary of <Topic>. <Topic> should be a topic that was never discussed within the meeting. \\
\small 2. The question should be a rephrase of asking what <Speaker> said regarding to <Topic> in the meeting. <Topic> should be a topic that was never discussed by <Speaker> within the meeting.\\
\small 3. The question should be a rephrase of asking what <Speaker> said regarding <Topic> in the meeting. <Speaker> should be a name of a person that did not participate in the meeting.\\
\small 4. The question should be a rephrase of asking what was the advantage of <Solution>. <Solution> should be something that was never discussed within the meeting.\\
\small 5. The question should be a rephrase of asking what was decided regarding <Topic>. <Topic> should be a topic that was never discussed within the meeting.\\
\small 6. The question should be a rephrase of asking what was decided regarding <Topic>. <Topic> should be a topic that was discussed within the meeting but never got to any conclusion.\\
\hline
\small \textbf{Context-dependent query templates} \\
\small 1. The question should be an implicit follow-up question regarding the previous message in the dialog using demonstrative pronouns such as "It", "He", "She", "They", "That". Example: "What did he say about it?", "What was their conclusion?"\\
\small 2. The question should include a word with the same meaning as "else"/"other"/"besides".\\
\hline
\end{tabular}
\caption{All query templates. The bracketed placeholders (e.g., <Topic>) signal where relevant meeting-specific topics should be inserted. Empirical evidence indicates that the LLM demonstrates a capacity to interpret these bracketed instructions, successfully generating contextually appropriate queries.}
\label{full_query_templates_list}
\end{table*}

\begin{table*}[ht]
\centering
\begin{tabular}{p{6in}}
\hline \\
\textbf{Agent response full prompt} \\
\textit{<Meeting transcript>} \\
\textit{<Dialog history>}\\
\textit{<User query>}\\
Instructions: The response should answer the user's question based on the meeting. \\
The response should follow the rules: \\
\begin{itemize}
    \item The response should be always directly derived from the meeting. It should not include any opinion or fact that was not presented within the meeting.
    \item The response should take one of two forms:
    \begin{enumerate}
        \item Up to 3 sentences of free text.
        \item Up to 2 intro sentence and between 3 and 5 bullet points. Mark each bullet point with a single asterisk ('*'). for example:\newline
          "Here are the topics discussed in the meeting:\newline
          * <Topic1>\newline
          * <Topic2>\newline
          * <Topic3>",\newline
          "The participants raised few concerns regarding the timeline:\newline
          * <Concern1>\newline
          * <Concern2> \newline
          * <Concern3>"\newline
        Choose the form that best suit the answer.
    \end{enumerate}
    \item When referencing the participants of the meeting as a group, the response should refer to them as "The participants".
    \item When referencing to a single person, use gender neutral pronouns such as "They" or "Them".
    \item When referencing to the meeting, refer to it as "the meeting" and NOT "the meeting transcript".
\end{itemize} \\
\hline
\end{tabular}
\caption{Full system's response prompts}
\label{table:response_prompt}
\end{table*}

\begin{table*}[ht]
\centering
\begin{tabular}{|p{0.23\linewidth} | p{0.70\linewidth}|}
\hline
\textit{<Meeting transcript>} &
T\#0 Grad C said: Nice. \newline
T\#1 Grad D said: OK. \newline
T\#2 Grad A said: to to handle. \newline
T\#3 Grad D said: Is that good? \newline
T\#4 Grad C said: Right. Yeah, I've have never handled them. \newline
...\\
\hline
\textit{<Instructions>} &
<Markers instructions> Generate the next response in the dialog between user and bot, adding a reference to the indices the answer comes from.\\
\hline    
\textit{<Dialog history>} &
user: What did the meeting participants decide to do to move the project forward?  \newline
bot: The participants decided to create a middle layer in their belief-net model...\\
\hline
\textit{<User query>} &
user: What did Grad C suggest to do when discussing the middle layer for the belief net model?\newline
bot: \\
\hline
\end{tabular}
\caption{Prompt for LLM agents at training and inference time.}
\label{table_inference_prompt}
\end{table*}

\begin{table*}[ht]
\centering
\begin{tabular}{|p{0.75\linewidth}|}
\hline
(T\#734,T\#759) Grad C suggested that values could be expanded... \\
\hline 
\end{tabular}
\caption{Example target format for fine-tuning. It includes the attribution spans and the expected response.}
\label{table_target}
\end{table*}

\begin{table*}[h]
\centering
\begin{tabular}{|lccc|c|}
\hline 
& \textbf{Train} & \textbf{Validation} & \textbf{Test} & \textbf{Overall} \\
\hline 
\# meetings & 157 & 34 & 34 & 225 \\
\# dialogs & 303 & 63 & 66 & 432\\ 
\# query-response pairs & 2922 & 611 & 628 & 4161 \\ 
\begin{tabular}{ll}
    & \begin{tabular}[t]{l} context-dependent
    \end{tabular} \\ 
\end{tabular} & 388 (13.28\%) & 81 (13.26\%) & 79 (12.58\%) & 548 (13.17\%) \\
\begin{tabular}{ll}
 & \begin{tabular}[t]{l} unanswerable
    \end{tabular} \\ 
\end{tabular} & 927 (31.72\%) & 149 (24.39\%) & 198 (31.53\%) & 1274 (30.62\%) \\
\begin{tabular}{ll}
    & \begin{tabular}[t]{l} query type: general
    \end{tabular} \\ 
\end{tabular} & 644 (22.04\%) & 88 (14.40\%) & 138 (21.97\%) & 870 (20.91\%) \\
\begin{tabular}{ll}
    & \begin{tabular}[t]{l} query type: specific
    \end{tabular} \\ 
\end{tabular} & 1515 (51.85\%) & 358 (58.27\%) & 308 (49.04\%) & 2179 (52.37\%) \\
\begin{tabular}{ll}
    & \begin{tabular}[t]{l} query type: yes/no
    \end{tabular} \\ 
\end{tabular} & 763 (26.11\%) & 167 (27.33\%) & 182 (28.98\%) & 1112 (26.72\%) \\
\hline 
\end{tabular}
\caption{Statistics of the MISeD dataset by split.}
\label{table:statistics_by_split}
\end{table*} 

\begin{figure*}[h!]
\centering
    \includegraphics[width=1\textwidth]{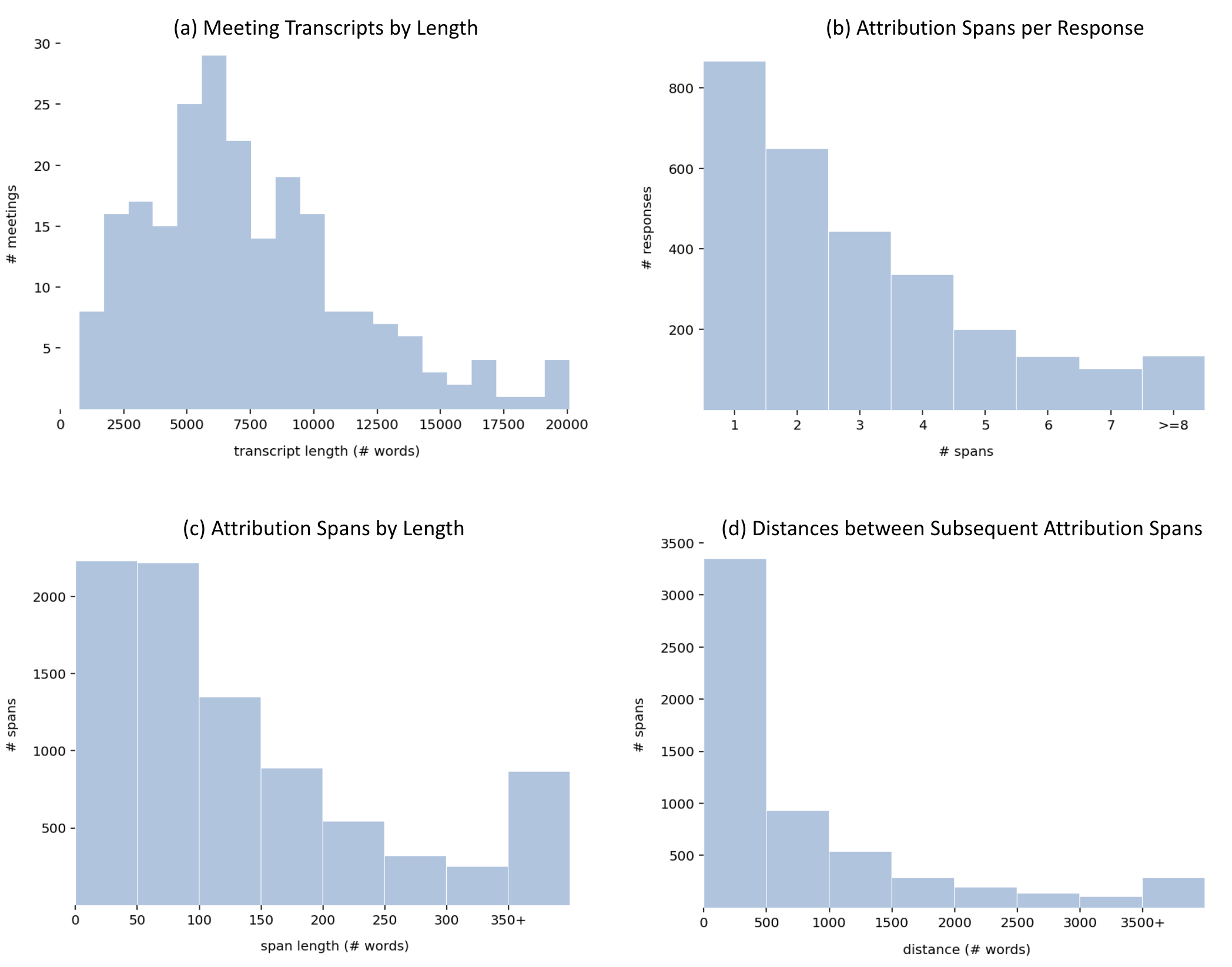}
    \caption{Transcript and attribution statistic. (a) Distribution of transcript length across the meetings used for MISeD. (b) Number of attribution spans in MISeD responses (among responses with attribution). (c) Distribution of attribution span length. (d) Distances between subsequent response attribution spans.}
    \label{fig:trancript_and_attributions}
\end{figure*}

\begin{table*}[h]
\centering
\begin{tabular}{|lcc|}
\hline 
& \textbf{WOZ} & \textbf{MISeD Test Set} \\
\hline 
\# meetings & 35 & 34 \\
\# dialogs & 70 & 66 \\ 
\# query-response pairs & 700 & 628 \\ 
\ overall vocabulary size & 2012 & 2492 \\ 
\ avg. response length & 28.52 & 41.71 \\ 
\ avg. response vocabulary size & 23.18 & 30.53 \\ 
\ avg. \% of response vocabulary overlapping with the transcript & 84.86\% & 78.43\% \\ 
\hline 
\end{tabular}
\caption{Statistics of the WOZ test set compared to the MISeD test set. MISeD responses tend to be longer, and to use a broader vocabulary, overlapping less with the transcript vocabulary. Vocabulary size is calculated as the number of unique lemmas.}
\label{table:woz_vs_mised}
\end{table*}

\begin{figure*}[h!]
\centering
    \includegraphics[width=1\textwidth]{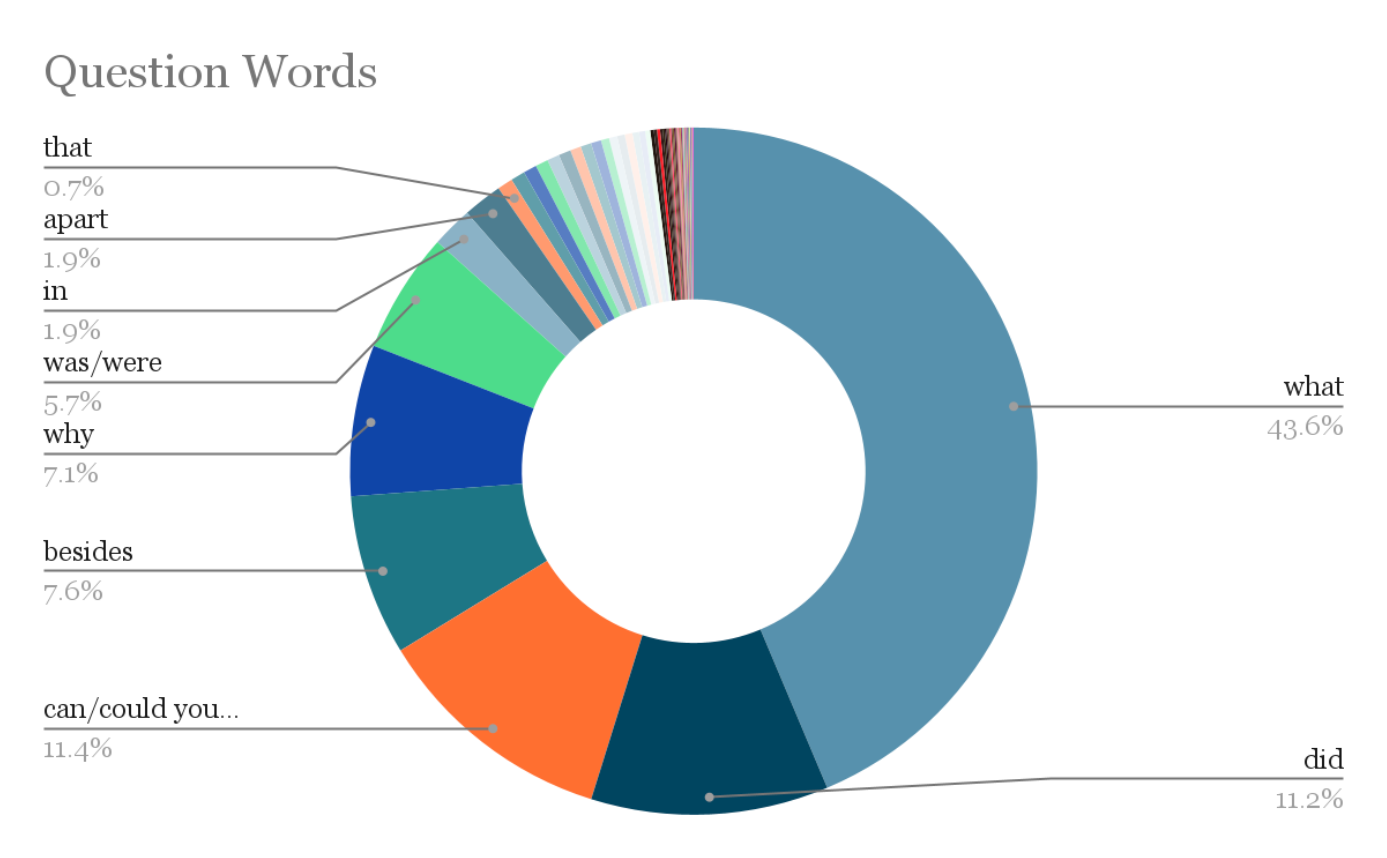}
    \caption{Distribution of question words (query prefix) within the dataset.}
    \label{fig:question_words}
\end{figure*}

\begin{table*}[ht]
\centering
\begin{tabular}{p{6in}}
\toprule 
\large\textbf{MISeD dialog example} \\
\\
\textbf{\textit{query:}} "Can you give me a summary of the group's battery life concerns?" \\
\textbf{\textit{answer:}} "The participants were concerned about the device's energy consumption, as it has a number of features that could quickly drain the battery. They discussed using both batteries and solar cells to power the device." \\
\textbf{\textit{relevant text spans:}} 55, 68, 140-149, 177 \\
\textbf{\textit{metadata:}} query type: general \\
\\
\textbf{\textit{query:}} "And what devices could it support?" \\
\textbf{\textit{answer:}} "The device is compatible with batteries and solar cells as a power source. The participants determined that the device could support an L\_C\_D\_ screen, an infra-red unit, and speech recognition." \\
\textbf{\textit{relevant text spans:}} 68, 140-141, 170-177, 266-270 \\
\textbf{\textit{metadata:}} query type: specific | context-dependent\\
\\
\textbf{\textit{query:}} "What was the advantage of using a push-button design on the remote?" \\
\textbf{\textit{answer:}} "The advantage of using a push-button design on the remote was not discussed in the meeting." \\
\textbf{\textit{metadata:}} query type: specific | unanswerable\\
\\
(... continue of the dialog) \\
\\
\textbf{\textit{query:}} "Did the participants agree to use a banana as the base model for the device's design?" \\
\textbf{\textit{answer:}} "Yes, the participants agreed to use a banana as the base model for the device\'s design. They felt that it would be a good way to incorporate the fruit and vegetable motif that was suggested by the fashion team."
 \\
\textbf{\textit{relevant text spans:}} 224-226 \\
\textbf{\textit{metadata:}} query type: yes-no \\
\\
\midrule 
\\
\large\textbf{QMSum query-based summarization examples} \\
\\
\textbf{General query:}\\
\textbf{\textit{query:}} "Summarize the whole meeting.", \\
\textbf{\textit{answer:}} "The meeting was mainly about the conceptual design of the remote control. Firstly, marketing introduced market research and fashion trends around the world. While the user interface emphasized that easy to use was more important. (...) After that, they discussed the shape of the remote control and decided to make it like a banana. Based on this shape, they talked about the detailed design of the appearance and user interface." \\
\\
\textbf{Specific query:} \\
\textbf{\textit{query:}} "Summarize the discussion about the result of marketing research and fashion trends." \\
\textbf{\textit{answer:}} "From the marketing's point of view, fancy was the highest priority, followed by technologically innovativeness and ease of use. The marketing expected the feel of the material to be spongy." \\
\textbf{\textit{relevant text spans:}} 17, 57 \\
\\
(... more specific standalone queries)
\\
\bottomrule
\end{tabular}
\caption{Example of a MISeD dialog vs. QMSum single-turn query-based summarization, over the same transcript. MISeD includes queries of additional types and characteristics, such as yes/no, unanswerable, and context-dependent queries.}
\label{mised_vs_qmsum_examples}
\end{table*}

\end{document}